\documentclass{article}
\usepackage{ijcai21}

\usepackage{times}
\usepackage{soul}
\usepackage{url}
\usepackage[hidelinks]{hyperref}
\usepackage[utf8]{inputenc}
\usepackage[small]{caption}
\usepackage{graphicx}
\usepackage{amsmath}
\usepackage{amsthm}
\usepackage{booktabs}
\usepackage{algorithm}
\usepackage{algorithmic}
\usepackage{flushend}

\usepackage{booktabs}       
\usepackage{amsfonts}       
\usepackage{nicefrac}       
\usepackage{microtype}      
\usepackage{multicol}
\usepackage{graphicx}
\usepackage{graphics}
\usepackage{multicol}
\usepackage{rotating}
\usepackage{mathrsfs} 
\usepackage{subcaption}
\usepackage{appendix}
\usepackage{array,multirow}
\usepackage{lscape} 
\usepackage{array}
\usepackage{sectsty}
\sectionfont{\fontsize{11}{12}\selectfont}
\usepackage{multicol}
\usepackage{xcolor}
\usepackage{colortbl}
\usepackage{todonotes}
\graphicspath{{figs/}}

\usepackage[export]{adjustbox}

\newcommand{\VIBR}{REF-VIB}

\title{Learning Robust Variational Information Bottleneck with Reference}

\author{
}

\author{
Weizhu Qian$^1$
\and
Bowei Chen$^2$\footnote{corresponding author} \and
Xiaowei Huang$^{3}$
\affiliations
$^1$Aalborg University\\
$^2$University of Glasgow\\
$^3$University of Liverpool
\emails
wqian@cs.aau.dk,
bowei.chen@glasgow.ac.uk,
xiaowei.huang@liverpool.ac.uk
}

\begin{document}

\maketitle

\begin{abstract}
We propose a new approach to train a variational information bottleneck (VIB) that improves its robustness to adversarial perturbations. Unlike the traditional methods where the hard labels are usually used for the classification task, we refine the categorical class information in the training phase with soft labels which are obtained from a pre-trained reference neural network and can reflect the likelihood of the original class labels. We also relax the Gaussian posterior assumption in the VIB implementation by using the mutual information neural estimation. Extensive experiments have been performed with the MNIST and CIFAR-10 datasets, and the results show that our proposed approach significantly outperforms the benchmarked models.    
\end{abstract}

\section{Introduction}


From the information-theoretical view of deep learning, the internal representation of some hidden layers is considered as a stochastic encoding $Z$ of the input $X$, and the goal of learning is to obtain an encoding that is maximally informative about the target $Y$. Therefore, information bottleneck (IB) approaches \cite{tishby2000information} force $Z$ to be a minimal latent representation of $X$ for predicting $Y$. 
The 
IB
methods have recently attracted significant attention in their ability of learning good hidden representations. See \cite{goldfeld2020information} for a recent survey. The 
IB methods
can also 
be used in sparse neural networks using Dropout \cite{achille2018information} and to compress neural networks \cite{dai2018compressing}. \cite{peng2019variational} developed a Variational Discriminator Bottleneck which can be applied to many applications, such as imitation learning, inverse reinforcement learning, and generative adversarial networks. 

Variational information bottleneck (VIB), started from \cite{alemi2017deep}, leverages variational inference to implement the IB method \cite{tishby2000information,tishby2015deep}. VIB models (or stochastic neural networks fit using VIB method), are believed to be more robust than deterministic models (e.g., conventional deep neural networks) because VIB learns the disentangled latent representations $Z$ which ignore the details of the input $X$ as many as possible \cite{alemi2017deep}. This has been improved by \cite{federici2020learning} with a multi-view approach where each view represents a task. 
Specifically, the robustness of learned representations can be improved by eliminating task-related information. However, additional information about multiple tasks is required.  
Moreover, \cite{belghazi2018mutual} leverages Mutual Information Neural Estimation to 
help with the implementation of VIB, without concerning  robustness performance. \cite{goyal2020variational} proposes Variational Bandwidth Bottleneck which compresses only part of the input, and shows that it has better generalization for reinforcement learning tasks. \cite{9088132} considers the robustness of VIB by adapting the variational bound of \cite{alemi2017deep}, but this method is only implemented via fully-connected layers and tested on FGSM attacks.

In this paper, we propose a novel method that learns robust VIB models that is radically different from the above methods (i.e., without taking the multi-view approach, is able to work with convolutional networks, and is robust against more involved attacks). The main idea is to use soft labels as the learning targets. The soft labels are pre-learned from a reference network. Specifically, they are the numerical outputs of the last hidden layer of the reference network but not the logits of the penultimate layer as in knowledge distillation \cite{hinton2015distilling,papernot2016distillation}. By using soft labels as training targets, the VIB training has been transformed from the original classification problem into a maximum likelihood estimation based regression problem so the outputs share the same manifold. Moreover, \cite{vincent2008extracting,vincent2010stacked} showed that an autoencoder structure trained with perturbed images as the input and the clean images as the target can be used to reconstruct \lq\lq{}repaired\rq\rq{} images from perturbed ones. Our method is inspired by this method to some extent, but with completely different purposes.

There are two major advantages of soft labels. On the one hand, they would lessen the confidence of the model and prevent it from overfitting~\cite{pmlr-v97-cohen19c}. On the other hand, they improve the efficiency of model training. It is widely acknowledged that model training with the single-step adversarial examples can easily lead to the risk of \emph{label leaking}~\cite{kurakin2016adversarial}, that is, the model's performance on adversarial examples is better than that with the clean examples. Therefore, many other studies employ iterative adversarial examples but the model training process becomes computationally expensive. Using the soft labels pre-learned from the reference network can help circumvent the risk of label leaking with the single-step adversarial examples, whose generation process such as FGSM \cite{goodfellow2014explaining} is much simpler and quicker.

Moreover, in the VIB, we replace the reparameterization method \cite{kingma2014auto} with the mutual information neural estimation (in short MINE) \cite{belghazi2018mutual}. The latter does not rely on Gaussian assumption for the posterior so it also improves the robustness of VIB. To validate our PROP method, extensive experiments are performed on MNIST and CIFAR-10 datasets. We examine both FGSM and PGD attacks at different scales, and compare the PROP method with several widely used methods such as adversarial training, knowledge distillation. The experimental results verify the robustness improvement of our method.

\section{Model}

Let $X$ be the input, $Z$ the latent representation of $X$, and $Y$ the target. Without loss of generality, we assume an information Markov chain $X \rightarrow Z \rightarrow Y$. The information bottleneck~\cite{tishby2000information,tishby2015deep} solves the following optimisation problem:
\begin{align}
\label{Eq:IB_1}
\text{max}  & ~ I(Z;Y),   \\
\label{Eq:IB_2}
\text{s.t.} & ~ I(Z;X) \leq I_C,
\end{align}
where $I$ is the mutual information and $I_C$ is a constant representing the information constraint. This is equivalent to maximizing the Lagrangian objective function
\begin{align}
\label{Eq:IB_Lag}
\mathscr{L}_{\text{IB}} = I(Z;Y) - \beta \big( I(Z;X) - I_C \big).
\end{align}
Based on Eq.~(\ref{Eq:IB_Lag}), the VIB model~\cite{alemi2017deep} computes $I(Z;X)$ via an encoder neural network and computes $I(Z;Y)$ via a classifier neural network, respectively. Our study aims to enhance the VIB model's robustness in face of adversarial attacks such as the projected gradient
descent (PGD) attack \cite{kurakin2016adversarial} and the fast gradient sign method (FGSM) attack \cite{goodfellow2014explaining}. 


\begin{figure}[t]
\centering
\includegraphics[width=1\linewidth]{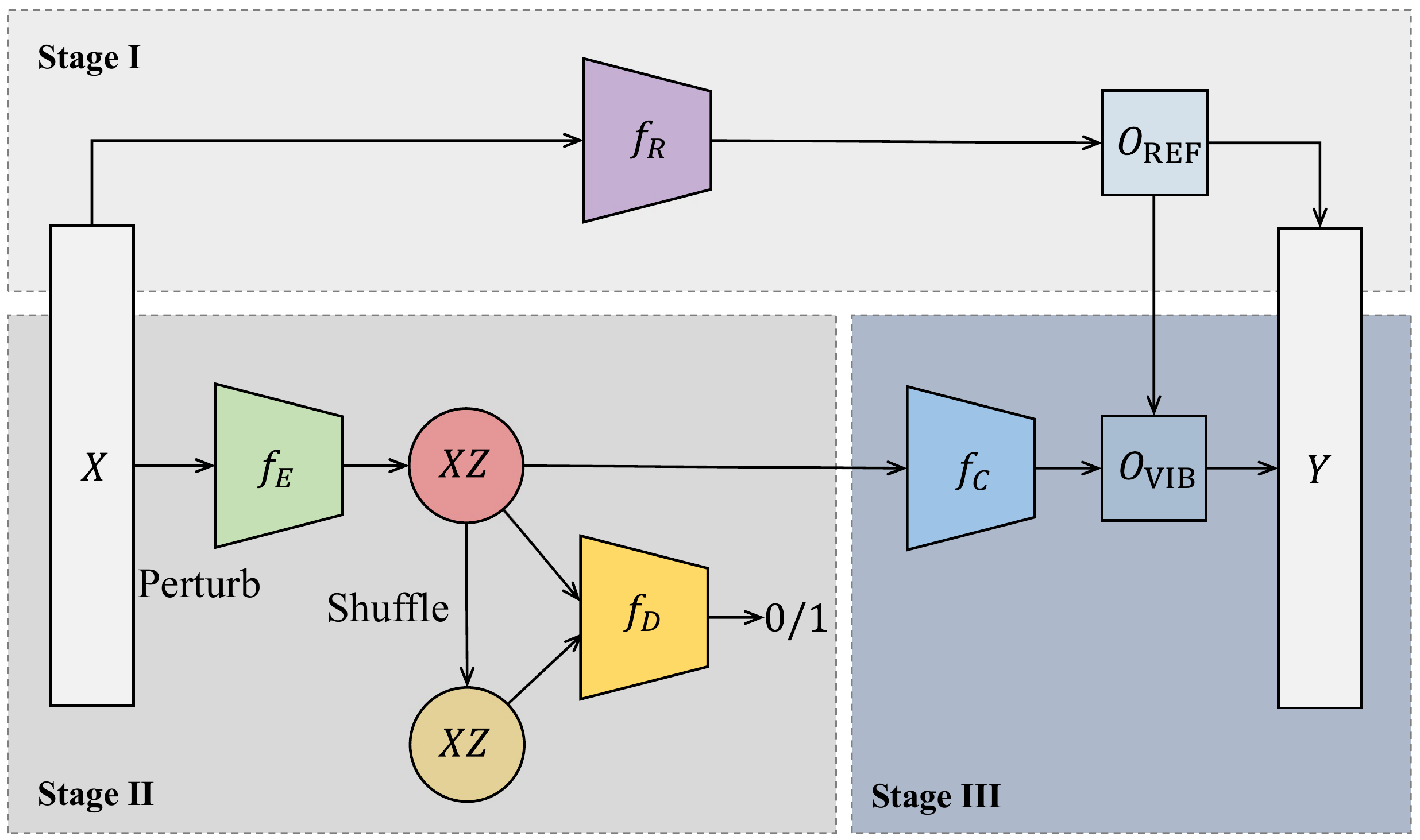}
\caption{Overview of our PROP robust learning method for the VIB, where $X$ is the input, $Y$ is the output, $f_E$ is the encoder network, $XZ$ (red) represents the joint samples, $XZ$ (yellow) represents the independent samples, $f_D$ is the discriminator, $f_R$ is the reference network, $f_C$ is the classifier, and $o_{\text{REF}}$ and $o_{\text{VIB}}$ are the soft labels from the reference network and the classifier, respectively.}
\label{fig:framework}
\end{figure}


The overview of our proposed method is presented in Fig.\ref{fig:framework}. Our model has a reference network $f_R$, an encoder $f_E$, a discriminator $f_D$ and a classifier $f_C$. It consists of three stages. In Stage I, a reference network $f_{R}(x)$ is pre-trained with clean samples $x$ to learn the soft labels $o_{\text{REF}}$, which are the output of the last hidden layer. The Stage II aims to compute the mutual information loss $I(Z;X)$ using the Mutual Information Neural Estimation (MINE) method~\cite{belghazi2018mutual}. The input $x'$ is firstly perturbed with either random noise (normal training) or adversarial attacks (adversarial training) to train an encoder network $f_E$. Then, the joint samples $XZ_\text{Joint}$ obtained by the encoder $f_E$ are shuffled to obtain the independent samples $XZ_\text{Ind}$. Afterwards, the discriminator $f_D$ is used to discriminate $XZ_\text{Joint}$ from $XZ_\text{Ind}$. The purpose of using $XZ_\text{Ind}$ and $f_D$ is due to our adoption of the Donsker-Varadhan representation of the Kullback–Leibler (KL) divergence. In Stage III, the learned soft labels are leveraged to compute the mean squared error loss $||o_{\text{VIB}}-o_{\text{REF}}||^2$ as the approximation of $I(Z;Y)$, where $o_{\text{VIB}}$ is the output of the last hidden layer of the VIB classifier $f_C$. 

We jointly train the encoder $f_E$, the discriminator $f_D$ and the classifier $f_C$ with the loss function $\mathscr{L}_{\text{PROP}}$. First, for a set of inputs $x_i$, we define the following loss function for the VIB model with reference: 
\begin{align}
\label{Eq:RVIB_loss}
\mathscr{L} & =
\sum_{i}^{N} \bigg[ \Big\| f_R(x_i)- f_C \big (f_E(x'_i) \big) \Big\|^2  \nonumber  \\   
& \qquad  + \beta \Big( \mathop{{}\mathbb{E}}{_{P_{XZ}}}\big [f_D \big] - \log \mathop{\mathbb{E}}{_{_{P_{X}\otimes P_{Z}}}\big[e^{f_D} \big]} \Big) \bigg ] 
\end{align}
where $N$ is the size of mini batch, and $\beta$ is a non-negative Lagrangian multiplier determining the optimisation trade-off. Based on Eq. (\ref{Eq:RVIB_loss}), let $\mathscr{L}_{\text{Clean}}(x_i^{\text{Clean}})$ and $\mathscr{L}_{\text{Adv}}(x_i^{\text{Adv}})$ be the loss functions trained on clean samples and adversarial samples, respectively. Then, we can have the loss function $\mathscr{L}_{\text{PROP}}$ for the adversarial training:
\begin{align}
\label{Eq:Final_loss}
\mathscr{L}_{\text{PROP}} & = \frac{\sum_{i \in \text{Clean}}{\mathscr{L}_{\text{Clean}}(x_i^{\text{Clean}})}  +\lambda \sum_{i \in \text{Adv} }{\mathscr{L}_{\text{Adv}}(x_i^{\text{Adv}})}}{m-k+\lambda k},
\end{align}
where $m$ is the total batch size, $k$ is the adversarial sample size, $\lambda$ is the relative weight. We call our new VIB method \VIBR\ to highlight the distinct feature of pre-learned soft labels. Algorithm~\ref{Alg:Algorithm} presents the overall training procedure. In the following, we explain the components of our \VIBR\ method with details.

\begin{algorithm}[tb]
\caption{REF-VIB training}
\label{Alg:Algorithm}
\textbf{Requirements}: $x$ (images) , $y$ (labels)

\begin{algorithmic}[1] 
\WHILE{ $\text{epoch} \leq \text{max epoch}$}
    \STATE{train a reference neural network $f_R(x)$;} 
\ENDWHILE

\WHILE{ $\text{epoch} \leq \text{max epoch}$} 

    \STATE{If $i$ $\in$ Clean:} 
        \STATE{\quad $x'_i$ $\gets$ $x_i$ + $\epsilon$}  
    \STATE{If $i$ $\in$ Adversarial:} 
        \STATE{\quad $x'_i$ $\gets$ attack $x_i$}  
            
        \STATE{$xz_{\text{Joint}}$ $\gets$ $f_E(x'_i)$}  
        
        \STATE{$xz_{\text{Ind}}$ $\gets$ shuffle $xz_{\text{Joint}}$ along the batch axis}   
        
        \STATE{$I(Z;X)$ $\gets$ $\mathop{{}\mathbb{E}}{_{P_{XZ}}}[f_D(xz_{\text{Joint}})]$ \\  
        \qquad \qquad \qquad $ - \log\{ \mathop{\mathbb{E}}{_{_{P_{X}\otimes P_{Z}}}[e^{f_D(xz_{\text{Ind}})}]} \}$} 
        
        
        
        \STATE{$I(Z;Y)$ $\gets$ $\|f_C(xz_{\text{Joint}})-f_R(x_i)\|^2$}
        
        \STATE{Minimize final loss function $\mathscr{L}_{\text{PROP}}$} 

\ENDWHILE

\end{algorithmic}
\end{algorithm}

The VIB model~\cite{alemi2017deep} implements Eq. (\ref{Eq:IB_Lag}) with reparameterization method \cite{kingma2014auto} via deep neural networks. 
%
%
It approximates $I(Z;X)$ and $I(Z;Y)$ with an encoder network and a decoder network, respectively, 
by deriving the variational bounds for them. 
The upper bound of $I(Z;X)$ yields
\begin{align}
\label{Eq:I_ZX_bound}
I(Z;X) \leq \mathbb{E}_x \left[D_\text{KL} \big(p(z|x)||q(z)\big)\right],    
\end{align}
where $D_\text{KL}$ is the Kullback-Leibler (KL) divergence, $q(z)$ is an uninformative prior distribution.
Similarly, the lower bound of $I(Z;Y)$ can be computed by
\begin{align}
\label{Eq:I_ZY_bound}
I(Z;Y) \geq & \mathbb{E}_{p(y,z)} \big [ \log \left\{ p(y|z) \right\} \big].   
\end{align}
Combining Eq. (\ref{Eq:I_ZX_bound}) and Eq. (\ref{Eq:I_ZY_bound}), the loss function becomes 
\begin{align}
\label{Eq:OP_VIB_Repa}
 \mathscr{L}_{\text{VIB}}  
 = &  \ 
 \mathbb{E}_{\mathscr{D}} \bigg[ \mathbb{E}_{z \sim p(z|x)} \Big[-\log\{ p(y|z)\} \Big] \nonumber \\
 & \hspace*{50pt} + \beta D_\text{KL}\big(p(z|x)|| q(z)\big ) \bigg],
\end{align}
where $p(z|x)$ is the encoder and $p(y|z)$ is the classifier.

In the following, we introduce how do we update Eqs. (\ref{Eq:I_ZX_bound})-(\ref{Eq:I_ZY_bound}) by taking the MINE method and the reference network, respectively. This will lead to the update of $\mathscr{L}_{\text{VIB}}$ into $\mathscr{L}$ of Eq. (\ref{Eq:RVIB_loss}), and VIB into the proposed \VIBR\ method.

\paragraph{Mutual Information Neural Estimation}

As for estimating the mutual information $I(X;Z)$, apart from using the re-parameterization method, we can alternatively use the Mutual Information Neural Estimation (MINE) method~\cite{belghazi2018mutual}. By using the MINE method, the Gaussian assumption on the posterior $p(z|x)$ can be relaxed, so the approximation of the true posterior $p(z|x)$ can be more accurate. 

First, we introduce the Donsker-Varadhan representation of the KL divergence \cite{donsker1983asymptotic,belghazi2018mutual}, defined as below
\begin{align}
\label{Eq:DV}
D_\text{KL}(P||Q) = \underset{T:\Omega \rightarrow \mathop{{}\mathbb{R}}}{\mathrm{sup}} \mathop{{}\mathbb{E}}{_{P}}\big[T\big] - \log \mathop{\mathbb{E}}{_{Q}\big[e^T\big]},
\end{align}
where $P$ is the posterior distribution, $Q$ is the prior distribution and $T$ is a discriminator network used to distinguish joint samples from independent samples. This can be implemented via a discriminator network using both joint and independent samples, as presented in the Stage II of Fig.~\ref{fig:framework}.  

The reason why we consider the KL divergence is because $I(X;Z)$ can be defined as follow: 
\begin{align}
\label{Eq:I_ZX}
I(Z;X) = & \ D_\text{KL} \big (\mathop{{}\mathbb{{P_{XZ}}}}||\mathop{{}\mathbb{{P_{X}}}} \otimes \mathop{{}\mathbb{{P_{Z}}}} \big ) 
\end{align}
where $\mathbb{{P_{XZ}}}$ is the joint distribution, $\mathbb{{P_{X}}}$ and $\mathbb{{P_{Z}}}$ are the marginal distributions.
Therefore, as opposed to the variational upper bound in Eq. (\ref{Eq:I_ZX_bound}), we use bound $I(Z;X) \geq I_\theta(Z;X)$ to estimate $I(Z;X)$ according to Eq. (\ref{Eq:DV}), such that 
\begin{equation}
\label{Eq:I_ZX_DV}
I_\theta(Z;X) = \underset{\theta \in \Theta}{\mathrm{sup}} \mathop{{}\mathbb{E}}{_{P_{XZ}}}\big[T_\theta\big] - \log \mathop{\mathbb{E}}{_{_{P_{X}\otimes P_{Z}}}\big[e^{T_\theta}\big]},
\end{equation}
where $T_\theta$ is a family of functions parametrized by a discriminator network (i.e., $f_D$ shown in Fig.~\ref{fig:framework}) with parameters $\theta \in \Theta$.  Hereafter, we omit $\theta$ for simplicity. To solve Eq. (\ref{Eq:I_ZX_DV}), we use the mini batches of the input to feed an encoder network so as to acquire the samples of the joint distribution $P_{XZ}$ (positive samples). While, the samples of $P_{X}\otimes P_{Z}$ can be obtained through shuffling the samples of $P_{XZ}$ along the batch axis (negative samples).

To simplify the discussion, we use \emph{VIB-M} to denote the VIB model trained using the MINE method and \emph{VIB-R} to denote the VIB model trained through the re-parameterization method. Combining Eq. (\ref{Eq:I_ZY_bound}) and Eq. (\ref{Eq:I_ZX_DV}), the loss function of the VIB-M model can be obtained as follows: 
\begin{align}
\label{Eq:OP_VIB_Mine}
& \mathscr{L}_{\text{VIB-M}} =  \mathbb{E}_{\mathscr{f_D}} \bigg[ \mathbb{E}_{z \sim p(z|x)} \big[-\log\{ p(y|z)\} \big]  \nonumber \\
& \hspace*{45pt} + \beta \Big (\mathop{{}\mathbb{E}}{_{P_{XZ}}}\big[f_D\big] - \log \mathop{\mathbb{E}}{_{_{P_{X}\otimes P_{Z}}}\big[e^{f_D}\big]} \Big ) \bigg]. 
\end{align}

It should be noted that $\mathscr{L}_{\text{VIB-M}}$ is different from $\mathscr{L}$ in Eq. (\ref{Eq:RVIB_loss}) because of the term $\mathbb{E}_{z \sim p(z|x)} \big[-\log\{ p(y|z)\} \big]  \nonumber$, which will be replaced with $\|f_R(x_i)- f_C \big (f_E(x'_i) \big)\|^2 $ after using the reference network, discussed next.

\paragraph{Reference Network}
\label{sec:referencenetwork}

Another key novelty of our method is that we use a reference network to pre-learn the soft labels on clean samples, and then use the pre-learned  soft labels -- instead of original hard labels -- as the learning targets in the VIB model. These soft labels can be regarded as a set of numerical vectors corresponding to each input image in space. Note that the soft labels used in our method are s different from the labels used in Knowledge Distillation \cite{hinton2015distilling,mishra2017apprentice}, which are logits.  

The reason of doing this is that hard labels only retain the discrete categorical information of each input but discard the more detailed information, which may impede the learning performance of VIB. Moreover, with the guidance of the soft labels learned on clean samples, the predictions of VIB model on perturbed samples are projected onto the correct manifold during the training process.  

Now, according to Eq. (\ref{Eq:I_ZY_bound}), we use the pre-learned soft labels to compute the likelihood $p(y|z)$ to approximate the lower bound of $I(Z;Y)$. To this end, we let $f_R(x)$ be the output of the last hidden layer of the reference network and $f_C(xz_{\text{Joint}})=f_C \big (f_E(x'_i) \big)$ be the output of the last hidden layer of the classifier in the VIB. The distance between $f_R(\cdot)$ and $f_C(\cdot)$ can be measured by the Euclidean distance:
\begin{equation}
\label{Eq:Ref_loss}
\text{DIST}(o_{\text{REF}}, o_{\text{VIB}}) = \Big\|f_R(x_i)- f_C \big (f_E(x'_i) \big) \Big\|^2  
\end{equation}

Then, Eq. (\ref{Eq:Ref_loss}) is the loss function of the downstream task in the VIB, and by replacing $\mathbb{E}_{z \sim p(z|x)} \big[-\log\{ p(y|z)\} \big]  \nonumber$ of Eq. (\ref{Eq:OP_VIB_Mine}) with it, we obtain $\mathscr{L}$ of Eq. (\ref{Eq:RVIB_loss}). 

\paragraph{Adversarial Training with Reference}

Deep neural networks (DNNs) are known to be vulnerable to adversarial attacks. Even small perturbations on the clean images may significantly decrease the test performance of DNNs. The VIB model has been argued to be more robust than conventional neural networks \cite{alemi2017deep}. 
Nevertheless, as shown in Section~\ref{sec:experiments}, the VIB models are still not robust enough to withstand strong adversarial attacks, such as PGD attacks. See the comparison between the VIB-M (REF) and the VIB-M (REF+FGSM) models. Therefore, we consider another effective way of defending attacks, i.e., adversarial training, which uses adversarial samples as the training input. 

In this paper, we consider two popular gradient-based methods to generate adversarial samples, i.e., the FGSM attack~\cite{goodfellow2014explaining} and the PGD attack~\cite{kurakin2016adversarial}. Simply speaking, the FGSM attack is a single-step attack method and the PGD attack finds adversarial samples in an iterative way.








Let $x$ denote the clean samples and $x_\text{Adv}$ the adversarial samples. We can obtain the adversarial samples as follows:
\begin{align}
\label{Eq:Adv}
x^\text{Adv} = x + \delta
\end{align}
where $\delta$ is the perturbation. For any $x \in {X}$ with $l_p$-ball $\mathcal{B}(x, \epsilon)$ around $x: \{ x_\text{Adv} \in X: ||x_\text{Adv} - x_\text{Clean} || \leq \epsilon \}$ \cite{madaan2020adversarial}, in a VIB model, the related latent presentation $z$ is sampled from the input $x$ via an encoder and the $\hat{y}$ is corresponding prediction computed from $z$ via a classifier. This relation can be described as follow:  
\begin{align}
\label{Eq:Prediction}
\hat{y} =  f_C(z) =  f_C(f_E(x))
\end{align}
where $f_E(\cdot)$ is the encoder and $f_C(\cdot)$ is the classifier.

According to Eq. (\ref{Eq:Ref_loss}), to train the VIB classifier $f_C$ with adversarial samples, we minimize the following loss function:
\begin{align}
\label{Eq:Adv_ref}
    & \ \mathbb{E}_{\{\mathscr{D}, \delta\}} \Big[ \Big\|f_C(xz^\text{Adv}_{\text{Joint}})-f_R(x)\Big\|^2 \Big] \nonumber \\ 
=  & \ \mathbb{E}_{\{\mathscr{D}, \delta\}} \Big[ \Big\|f_C(f_E(x^\text{Adv}))-f_R(x)\Big\|^2 \Big],
\end{align}
where $xz^\text{Adv}_{\text{Joint}}$ is the joint latent variables learned by adversarial samples. We substitute the term $\mathbb{E}_{z \sim p(z|x)} \big[-\log\{ p(y|z)\} \big]$ in Eq. (\ref{Eq:OP_VIB_Mine}) with Eq. (\ref{Eq:Adv_ref}), then according to Eq. (\ref{Eq:RVIB_loss}), we have the adversarial training loss function, $\mathscr{L}_{\text{Adv}}$ in Eq. (\ref{Eq:Final_loss}) which is the final loss function of the PROP model. Eq. (\ref{Eq:Adv_ref}) suggests that we can learn the encoder and classifier jointly for the VIB model, combining adversarial training and the reference network. Consequently, the distortion of latent variables caused by adversarial attack can be suppressed as well, which leads to better generalization so as to defend attacks. 

The disadvantage of using PGD samples for training is its time consumption. One-step adversarial images, e.g., FGSM samples, are much faster to generate, however using such adversarial images to train conventional neural networks with hard labels will cause the \textbf{label leaking} effect \cite{kurakin2016adversarial}. Label leaking means the testing accuracy on adversarial images are much higher than the accuracy on clear images. This issue occurs because one-step adversarial images generated by simple transformations, and it is not hard for a deep neural network to learn such transformations, as a result, the deep neural network performs better on the adversarial images than the clean images. On the contrary, a more complex adversarial transformation, e.g., PGD, does not has this issue because its transformation is more difficult to learn. Fortunately, our method accelerates the learning process by leveraging FGSM samples without suffering from label leaking effect, as we use soft labels instead of hard labels.

\paragraph{Discussion}

We would like to discuss the rationale behind our selection of the three components in the proposed model.

\emph{Why Using the MINE Method?} We want to compute the posterior $p(z|x)$ as accurate as possible. Since the prior $q(z)$ is assumed to follow Gaussian, the reparameterization method assumes $p(z|x)$ follows Gaussian as well. However, the posterior can be more complex than the prior, and using a Gaussian distribution may not be accurate enough to approximate the prior. In the MINE method, we do not generate the posterior samples from a Gaussian distribution, instead, we generate the joint posterior samples with input perturbed by noise and use a discriminator to discriminate them from the shuffled samples. Once the model is trained, we can obtain the posterior samples with more complex shapes.            

\emph{Why Need Reference Network?} In a VIB model, the input is compressed to a latent distribution by the encoder. We believe that the vulnerability of a VIB is caused by the distortion of the latent distribution. Therefore, to improve the robustness VIB, we need to minimise the distortion (a similar idea was PROP in \cite{madaan2020adversarial}, but their method is based on Bayesian neural networks and pruning). The latent distribution of VIB is Gaussian-like, and the pre-learned soft labels are numerical vectors which contain more information than the original hard labels. Hence, to suppress the distortion of the latent distribution, using the soft labels to minimize the loss function of VIB is more effusive than using the hard labels. 

\emph{Why Performing Adversarial Training?} Adversarial training is known for improving the robustness of neural networks. The soft labels pre-learned by the reference network on the clean samples and the soft labels learned by the VIB model on the corrupted samples should share the same manifold. Consequently, we can leverage the pre-learned soft labels to guide the adversarial learning process of the VIB model to achieve a better performance. Our learning strategy is similar to denoising autoencoders \cite{vincent2008extracting,vincent2010stacked} to some extent. In denoising autoencoders, the corrupted images are used to train an autoencoder in order to recover 
clean images, which is an unsupervised learning scheme. Our method is supervised, and we use adversarial samples as the corrupted samples and soft labels as the target instead of original clean images.

\begin{table}[t]
\centering
\small
\caption{Summary of the examined models.}
\label{Tab:model_summary}
\centering
\begin{tabular}{|m{1.1in}|m{1.9in}|} 
\hline
Model & Description\\ 
\hline
Baseline & For the MNIST dataset, the baseline model architecture is CONV(64,3,1) + CONV(64,3,1) + FC(512) + FC(512) + FC(10); For the CIFAR-10 dataset, the baseline model is VGG16 \cite{simonyan2014very}. \\%
\hline
Baseline (KD) & The baseline model trained with KD \cite{papernot2016distillation}.\\%
\hline
Baseline (PGD) & The baseline model trained with PGD samples \cite{kurakin2016adversarial}. \\%
\hline
Baseline (REF) & The baseline model trained with a reference network.\\%
\hline
Baseline (PGD+REF) & The baseline model trained with trained with a reference network and PGD samples.\\%
\hline
VIB-R & The VIB model trained using the re-parameterization method~\cite{alemi2017deep}.\\%
\hline
VIB-M & The VIB model trained using the MINE method~\cite{belghazi2018mutual}.\\%
\hline
VIB-R (REF) & The VIB-R model trained with a reference network.\\%
\hline
VIB-M (REF) & The VIB-M model trained with a reference network.\\ %
\hline
VIB-M (PGD) & The VIB-M model trained with PGD samples.\\ %
\hline
VIB-M (REF+FGSM) & The VIB-M model trained with a reference network and FGSM samples.\\   
\hline
VIB-M (REF+PGD) &  The VIB-M model trained with a reference network and PGD samples.\\
\hline
\end{tabular}
\end{table}

\begin{table}[t]
\small
\centering
\caption{Model training settings, where $T$ is the temperature, $\epsilon$ is the magnitude of perturbation, $\alpha$ is the attack step size, $t$ is the attack step number, and $\eta$ is the dimension of latent variables.}
\label{Tab:train_method}
\begin{tabular}{|m{1.0in}|m{2in}|} 
\hline
Model & Description\\ 
\hline
KD & $T=1$.\\%
\hline 
PGD (MNIST) & $\alpha = 0.01$, $m=100$, $k=50$, $\lambda = 0.3$, $\epsilon = 0.3$, $t = 20$. \\%
\hline
FGSM (MNIST) & $\epsilon = 0.3$. \\
\hline
PGD (CIFAR-10) & $\alpha = 0.007$, $m=100$, $k=50$, $\lambda = 0.3$, $\epsilon = 0.03$, $t = 10$. \\%
\hline 
FGSM (CIFAR-10) & $\epsilon = 0.03$. \\
\hline
VIB-R & $\beta=0.001$, $\eta = 256$.\\%
\hline
VIB-M & $\beta=0.001$, $\eta = 256$.\\%
\hline
\end{tabular}
\vspace*{10pt}
\caption{Model testing settings, where $\epsilon$ is the magnitude of perturbation, $\alpha$ is the attack step size, $t$ is the attack step number.}
\label{Tab:test_method}
\centering
\begin{tabular}{|m{1.0in}|m{2in}|} 
\hline
Model & Description\\ 
\hline
FGSM (MNIST) &  $\epsilon = 0.05$, $0.10$, $0.15$, $0.20$, $0.25$, $0.30$. \\
\hline
PGD (MNIST) & $\epsilon = 0.05$, $0.10$, $0.15$, $0.20$, $0.25$, $0.30$, $\alpha = 0.01$, $t$ = $40$, random restart. \\%
\hline
FGSM (CIFAR-10) &  $\epsilon =0.01$, $0.02$, $0.03$, $0.04$, $0.05$, $0.06$.\\
\hline
PGD (CIFAR-10) & $\epsilon=0.01$, $0.02$, $0.03$, $0.04$, $0.05$, $0.06$, $\alpha = 0.007$, $t$ = $40$, random restart.  \\%
\hline
\end{tabular}
\end{table}

\section{Experiments}
\label{sec:experiments}

\paragraph{Datasets}

Two publicly available datasets are used in the experiments: (i) the MNIST dataset~\cite{lecun1998mnist} which includes $60, 000$ gray level handwritten images of size $28 \times 28$; and (ii) the CIFAR-10 dataset~\cite{krizhevsky2009learning} which includes $60, 000$ color images of size $32 \times 32$. For each dataset, we randomly sample $50, 000$ images for model training and $10, 000$ images for model testing. The sampled images from the CIFAR-10 dataset are normalized into $[0,1]^{32\times 32\times 3}$.

\paragraph{Experimental Settings}

To have a comprehensive comparison, different machine learning algorithms using various training methods are examined, including the recent advances like knowledge distillation (KD)~\cite{papernot2016distillation}, PGD attack \cite{kurakin2016adversarial} and FGSM attack \cite{goodfellow2014explaining}. Also, as the VIB model can be estimated by either the reparameterization method or the MINE method, both training methods are included in our experiments. To simplify the discussion, an algorithm using a specific training method is considered as a model here and it is denoted by a unique name. For the reader's convenience, we present all the examined models and their descriptions in Table~\ref{Tab:model_summary}. 

We adopt different settings of model architecture, training and testing. First, we use different neural networks as the baseline model and we have different architecture settings for the VIB model accordingly. For the MNIST dataset, the architecture of the baseline model is CONV(64,3,1) + CONV(64,3,1) + FC(512) + FC(512) + FC(10) and the architecture of the VIB model is CONV(32,3,1) + CONV(32,3,1) + FC(512) + FC(256) for encoder and FC(512) + FC(10) for classifier. For the CIFAR-10 dataset, the baseline models is VGG16 \cite{simonyan2014very} and the VIB model's architecture is CONV(64,3,1) + CONV(128,3,1) + CONV(256,3,1) + CONV(256,3,1) + CONV(512,3,1) +CONV(512,3,1) + FC(1024) + FC(512) + FC(256) for encoder and FC(512) + FC(1024) + FC(512) + FC(10) for classifier. Second, the baseline model and the VIB model are trained differently in these two datasets. Table~\ref{Tab:train_method} presents the details of model training settings. For the MNIST dataset, the optimizer is Adam \cite{kingma2014adam} and the learning rate is $0.001$. For the CIFAR-10 dataset, the optimizer is the stochastic gradient descent (SGD) with adaptive learning rate. Third, the trained models are evaluated under different attack settings in our datasets. As shown in Table~\ref{Tab:test_method}, the attack perturbation levels for the MNIST dataset range from 0.05 to 0.30 and are with the step size of 0.05 while they range from 0.01 to 0.06 and are with the step size of 0.01 for the CIFAR-10 dataset~\cite{kurakin2016adversarial}.

\paragraph{Results}

\begin{figure}[t]
\centering
\includegraphics[width=1\linewidth]{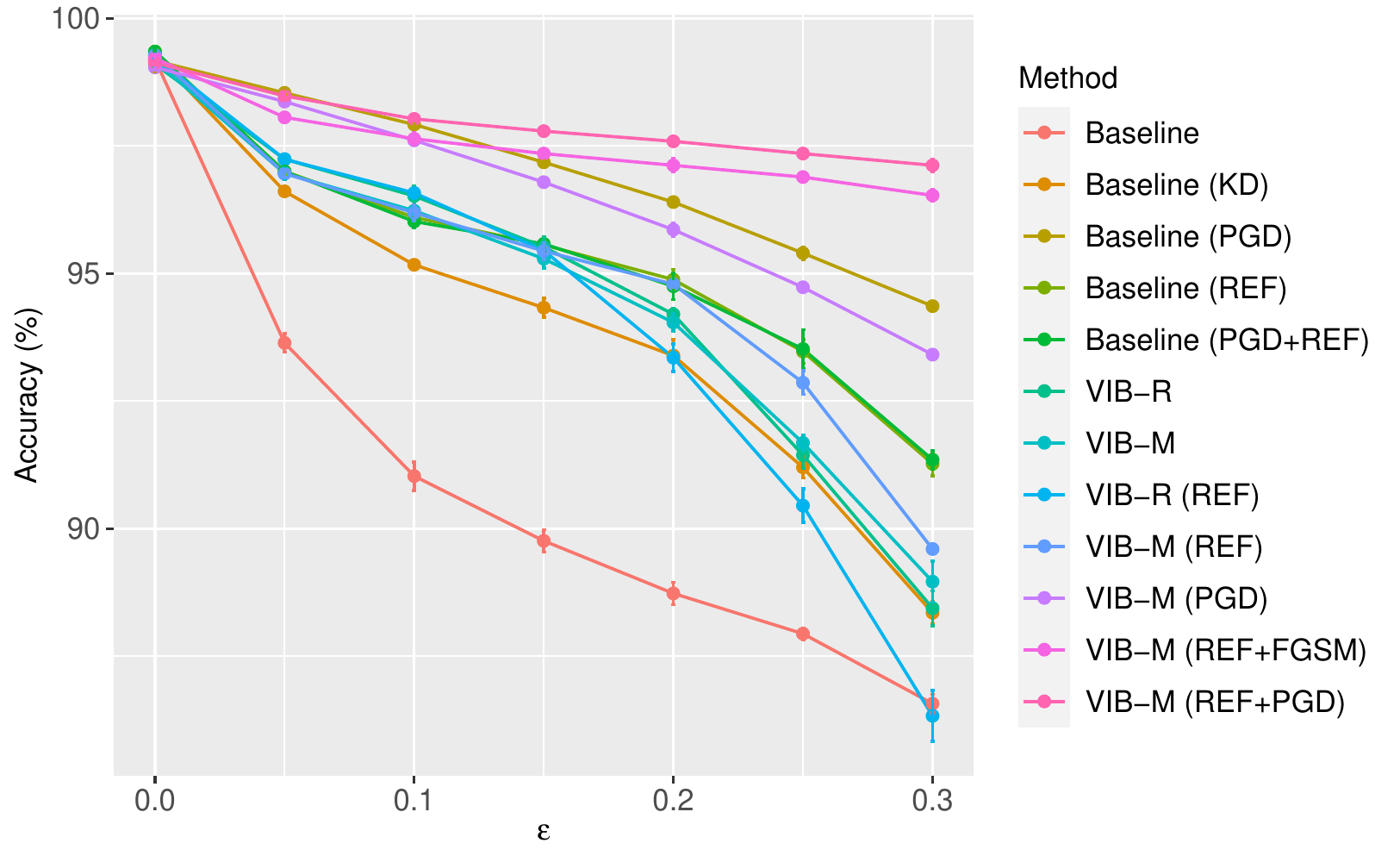} 
\vspace*{-12pt}
\caption{Models' classification test accuracy results ($\%$) with the FGSM attacks on the MNIST dataset.}
\label{fig:mnist_fgsm}
\vspace*{10pt}
\centering
\includegraphics[width=1\linewidth]{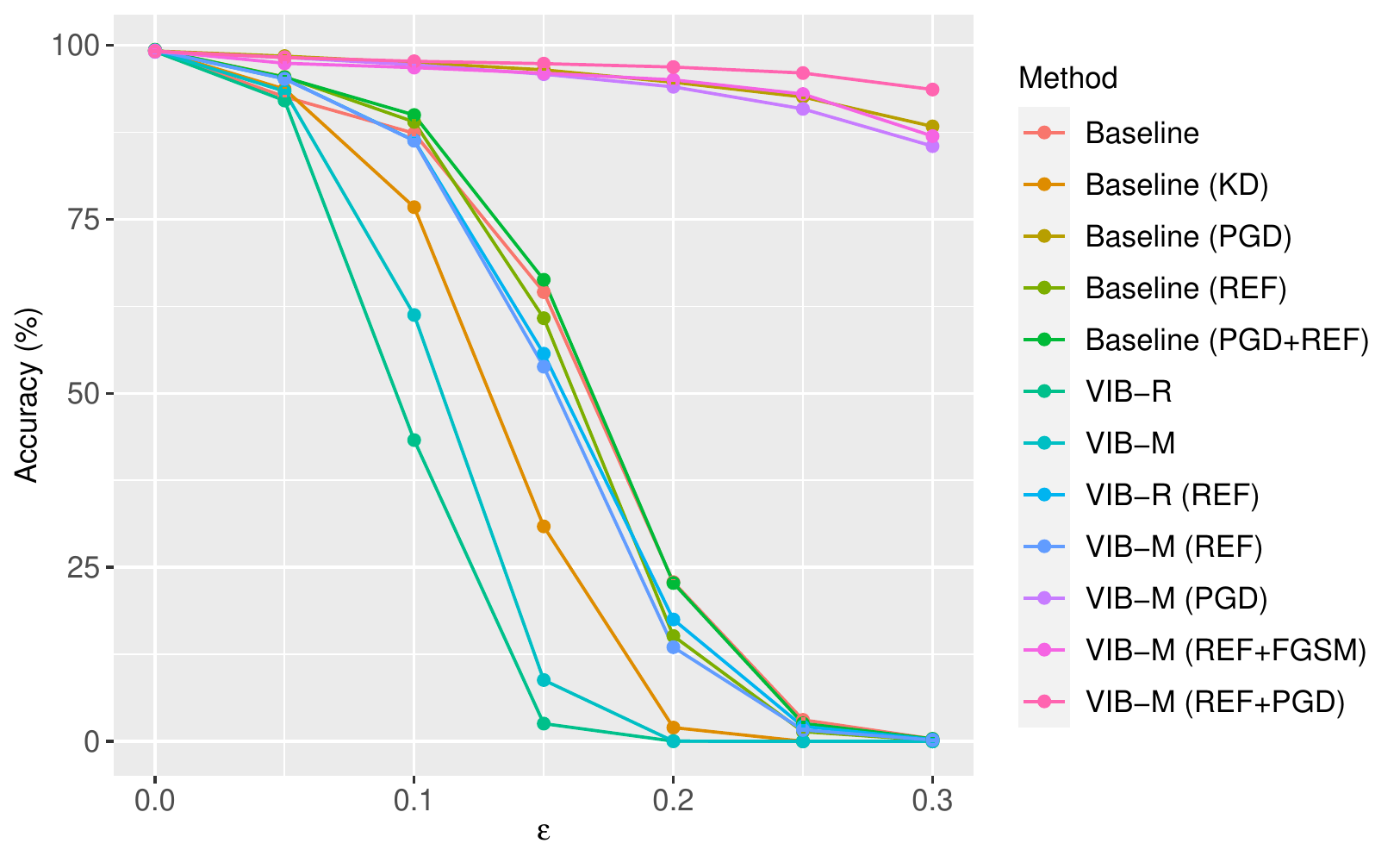}
\vspace*{-12pt}
\caption{Models' classification test accuracy results ($\%$) with the PGD attacks on the MNIST dataset.}
\label{fig:mnist_pgd}
\end{figure}

\begin{figure}[t]
\centering
\includegraphics[width=1\linewidth]{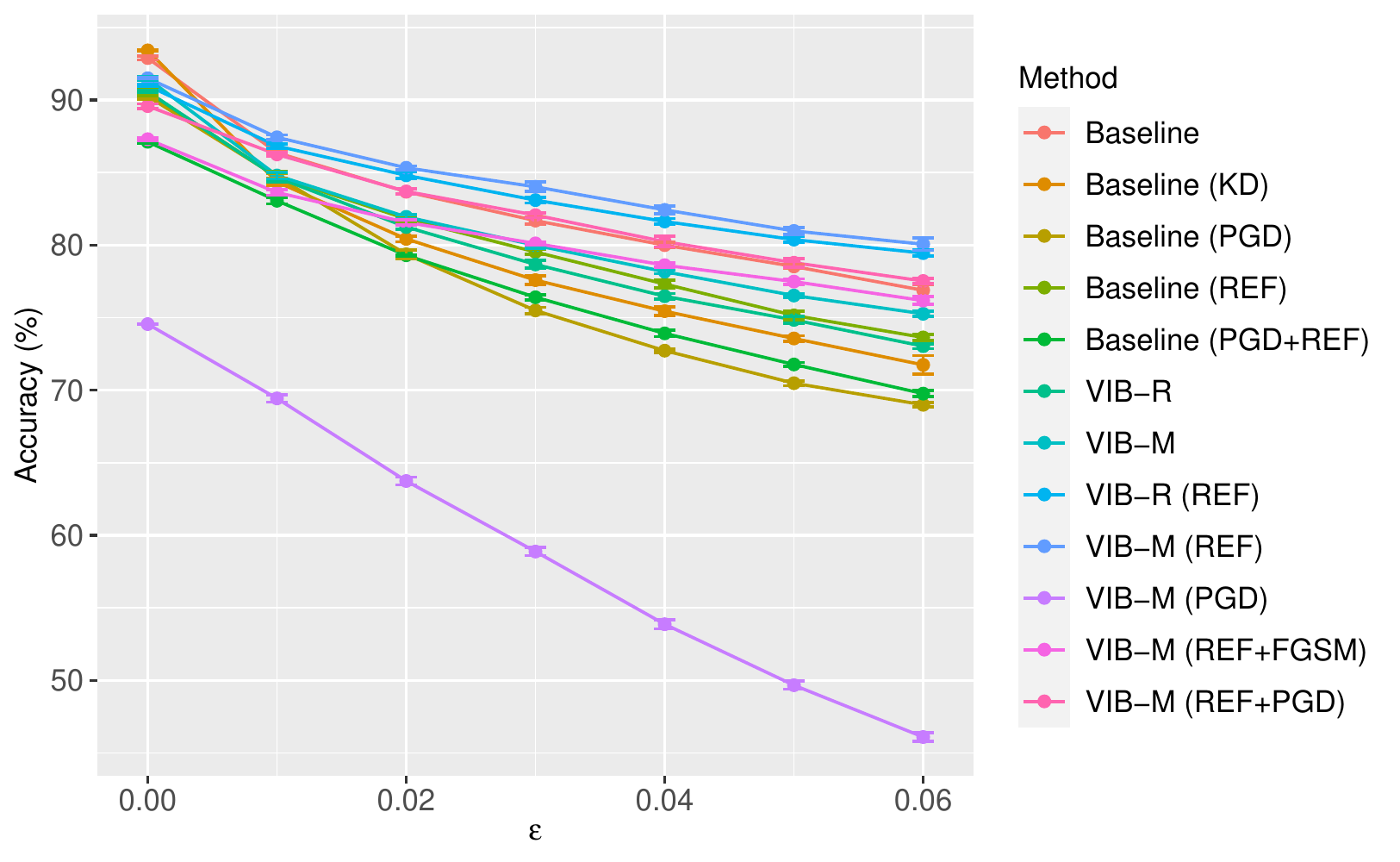}
\vspace*{-12pt}
\caption{Models' classification test accuracy results ($\%$) with the FGSM attacks on the CIFAR-10 dataset.}
\label{fig:cifar10_fgsm}
\vspace*{10pt}
\includegraphics[width=1\linewidth]{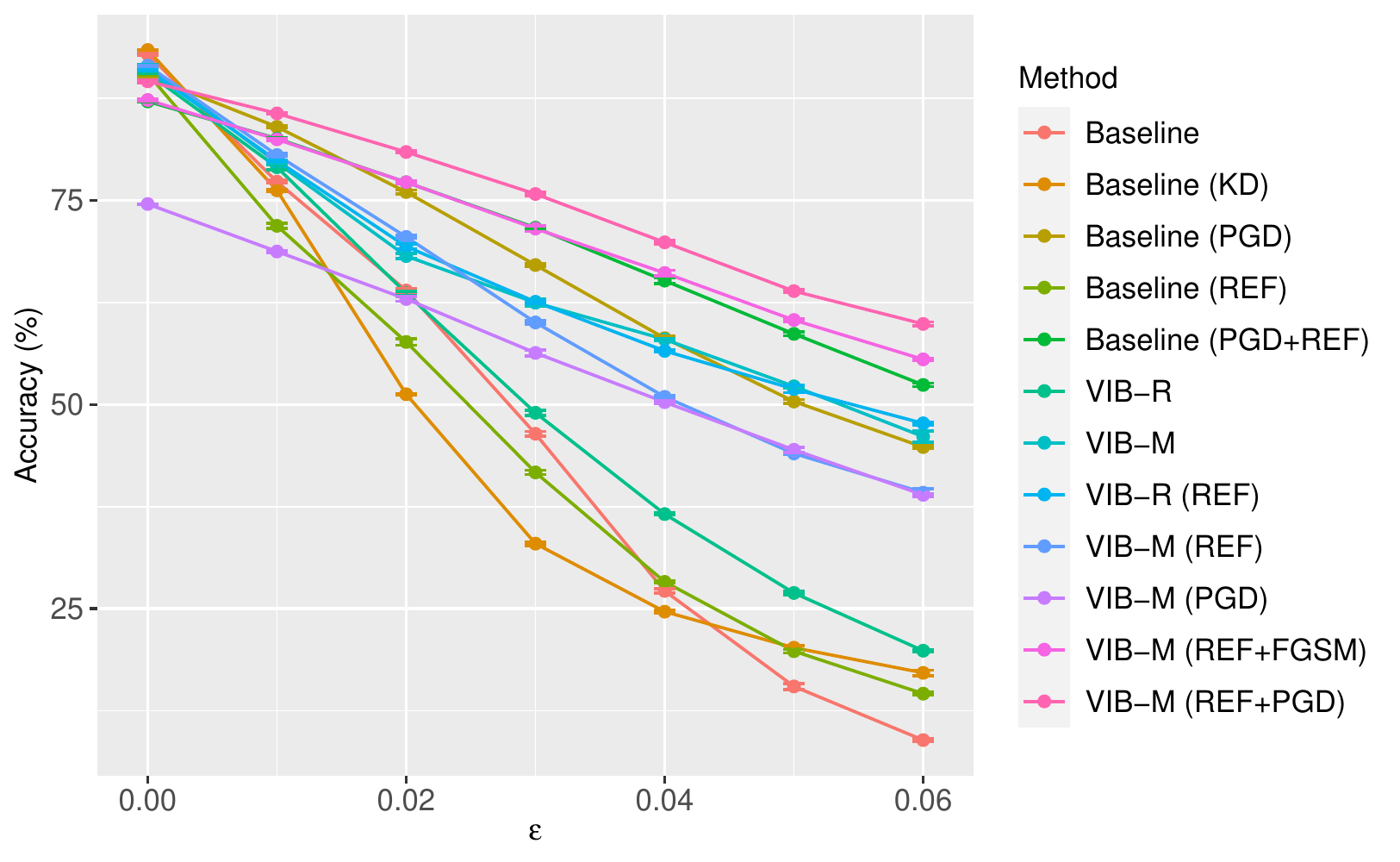}
\vspace*{-12pt}
\caption{Models' classification test accuracy results ($\%$) with the PGD attacks on the CIFAR-10 dataset. }
\label{fig:cifar10_pgd}
\end{figure}

We run 5 randomised trials for each examined model on the MNIST and the CIFAR-10 datasets, and report the models' classification results (i.e., the average accuracy and its standard deviation) of the test data in Figs.~\ref{fig:mnist_fgsm}-\ref{fig:cifar10_pgd}. 
Every curve represents a trained model, whose accuracy is gradually decreased with the increase of the parameter $\epsilon$ of either FGSM or PGD attack. A larger area under the curve (AUC) represents a more robust model. 

There are a few insights we would like to share.
%
First of all, the VIB model shows a good predictive capability. However, it is still vulnerable to strong attacks like the PGD attack. From the results on the MNIST dataset, we can see that the regularly-trained VIB models, i.e., the VIB-R and the VIB-M models, outperform the corresponding baseline models under the FGSM attacks. However, they become worse under the PGD attacks. Our experimental results also show that the MINE method generally outperforms the reparamterization method for training the VIB model because the former can provide more complex posteriors. 

It is generally believed that training a deep learning model with adversarial examples can improve the model's robustness, particularly, under the PGD attack. However, according to the MNIST results, the VIB (PGD) model does not achieve the best on any case. That indicates a naive adoption of adversarial training cannot improve the performance of the VIB model.  In fact, the results obtained by the VIB-M (REF+FGSM) and the VIB-M (REF+PGD) suggest only when the VIB model is trained with the support of reference network, the best performance can be achieved. \emph{By contrast, using our proposed methods, VIB-M(REF+FGSM/PGD), the robustness of the VIB models can be significantly improved under both FGSM and PGD attacks.} Between the VIB-M (REF+FGSM) and the VIB-M (REF+PGD) models, the former enjoys faster training while maintains comparable performance. From the results on CIFAR-10, we see that VIB-R/M do not necessarily outperform the baselines under either the FGSM or the PGD attacks. The VIB-R/M (REF) models have better performance under the FGSM attacks, but do not perform very well under the PGD attack. Our proposed methods, the VIB-M (REF+FGSM/PGD) models, have similar excellent performance against both FGSM and PDG attacks.

Overall, it can be seen that, first, learning with reference can improve the robustness of the VIB model; second, compared with other methods, adversarial training is more effective to defend the PGD attacks and using reference can improve the performance of adversarial training; third, as opposed to using hard labels, our method can use a single-step adversarial training method to accelerate the training process, i.e. the FGSM training, without causing the label leaking effect.

\section{Conclusion}

In this paper, we propose a novel method for training the VIB model which can produce accurate predictions as well as be robust under adversarial attacks. The idea is to train the VIB model by using the reference network and the MINE method. The former enables the mapping of both clean and perturbed inputs onto a smooth output manifold, so that the learned VIB model is smoother (and more robust). Also, this enables a fast training with FGSM-based method instead of PGD-based, without compromising the robustness performance. To the best of our knowledge, this is the very first study that integrates them into a VIB-based framework. The experimental results show that our proposed model outperforms the state-of-the-art benchmarked models.





\bibliographystyle{plain}
\bibliography{refs}

\pagebreak

\appendix

\begin{table*}[ht!]
\small
\caption{Models' classification test accuracy results ($\%$) with the FGSM attack on the MNIST dataset.}
\label{Tab:fgsm_mnist}
\centering
\begin{tabular}{|l|c|c|c|c|c|c|c|} \hline
 Method & Clean & $\epsilon=0.05$ & $\epsilon=0.1$ & $\epsilon=0.15$ & $\epsilon=0.2$ & $\epsilon=0.25$ & $\epsilon=0.3$ \\ \hline
 
 Baseline &$99.20_{\pm0.05}$ &$93.64_{\pm0.18}$ &$91.03_{\pm0.28}$ &$89.76_{\pm0.22}$ &$88.73_{\pm0.22}$ &$87.94_{\pm0.11}$ &$86.57_{\pm0.18}$\\%
 
 Baseline(KD) &$99.28_{\pm0.04}$ &$96.61_{\pm0.06}$ &$95.17_{\pm0.09}$ &$94.33_{\pm0.19}$ &$93.39_{\pm0.31}$ &$91.20_{\pm0.21}$ &$88.35_{\pm0.22}$ \\%
 
 Baseline(PGD) &$99.17_{\pm0.02}$ &\pmb{$98.54_{\pm0.03}$} &$97.92_{\pm0.08}$ &$97.18_{\pm0.06}$ &$96.40_{\pm0.09}$ &$95.40_{\pm0.12}$ &$94.36_{\pm0.06}$\\%
 
 Baseline(REF) &\pmb{$99.35_{\pm0.02}$} &$97.00_{\pm0.07}$ &$96.10_{\pm0.21}$ &$95.56_{\pm0.07}$ &$94.88_{\pm0.20}$ &$93.47_{\pm0.23}$ &$91.27_{\pm0.24}$\\%
 
 Baseline(PGD+REF) &\pmb{$99.35_{\pm0.03}$} &$97.00_{\pm0.11}$ &$96.02_{\pm0.12}$ &$95.58_{\pm0.14}$ &$94.75_{\pm0.26}$ &$93.52_{\pm0.38}$ &$91.35_{\pm0.18}$\\%
 
 VIB-R &$99.10_{\pm0.05}$ &$97.24_{\pm0.05}$ &$96.52_{\pm0.20}$ &$95.52_{\pm0.20}$ &$94.20_{\pm0.12}$ &$91.44_{\pm0.26}$ &$88.44_{\pm0.34}$ \\%
 
 VIB-M &$99.17_{\pm0.03}$ &$96.96_{\pm0.11}$ &$96.23_{\pm0.08}$ &$95.29_{\pm0.18}$ &$94.04_{\pm0.17}$ &$91.68_{\pm0.15}$ &$88.96_{\pm0.40}$\\%
 
 VIB-R(REF) &$99.24_{\pm0.03}$ &$97.24_{\pm0.07}$ &$96.58_{\pm0.10}$ &$95.45_{\pm0.12}$ &$93.35_{\pm0.27}$ &$90.45_{\pm0.33}$ &$86.33_{\pm0.50}$\\%
 
 VIB-M(REF) &$99.28_{\pm0.02}$ &$96.96_{\pm0.08}$ &$96.19_{\pm0.16}$ &$95.44_{\pm0.16}$ &$94.79_{\pm0.05}$ &$92.86_{\pm0.23}$ &$89.60_{\pm0.04}$\\ %
 
 VIB-M(PGD) &$99.04_{\pm0.05}$ &$98.37_{\pm0.04}$ &$97.61_{\pm0.08}$ &$96.79_{\pm0.08}$ &$95.86_{\pm0.13}$ &$94.73_{\pm0.06}$ &$93.41_{\pm0.10}$\\ %
 
 VIB-M(REF+FGSM) &$99.22_{\pm0.06}$ &$98.06_{\pm0.06}$ &$97.64_{\pm0.12}$ &$97.35_{\pm0.08}$ &$97.12_{\pm0.13}$ &$96.89_{\pm0.09}$ &$96.53_{\pm0.11}$\\   
 
 VIB-M(REF+PGD) &$99.13_{\pm0.04}$ &$98.48_{\pm0.08}$ &\pmb{$98.03_{\pm0.04}$} &\pmb{$97.79_{\pm0.07}$} &\pmb{$97.59_{\pm0.08}$} &\pmb{$97.35_{\pm0.05}$} &\pmb{$97.12_{\pm0.12}$} \\
 
 \hline
\end{tabular}
\end{table*}


\begin{table*}[ht!]
\small
\caption{Models' classification test accuracy results ($\%$) with the PGD attack accuracy on the MNIST dataset.}
\label{Tab:pgd_mnist}
\centering
\begin{tabular}{|l|c|c|c|c|c|c|c|} \hline
 Method & Clean &  $\epsilon=0.05$ & $\epsilon=0.1$ & $\epsilon=0.15$ & $\epsilon=0.2$ & $\epsilon=0.25$ & $\epsilon=0.3$ \\ \hline
 
 Baseline &$99.20_{\pm0.05}$ &$92.60_{\pm0.06}$ &$87.36_{\pm0.23}$ &$64.54_{\pm0.39}$ &$22.88_{\pm0.14}$ &$3.09_{\pm0.13}$ &$0.36_{\pm0.05}$ \\%
 
 Baseline(KD) &$99.28_{\pm0.04}$ &$93.71_{\pm0.10}$ &$76.74_{\pm0.13}$ &$30.88_{\pm0.12}$ &$1.98_{\pm0.08}$ &$0.01_{\pm0.01}$ &$0.00_{\pm0.00}$ \\%
 
 Baseline(PGD) &$99.17_{\pm0.02}$ &\pmb{$98.44_{\pm0.07}$} &$97.51_{\pm0.05}$ &$96.47_{\pm0.01}$ &$94.68_{\pm0.09}$ &$92.57_{\pm0.07}$ &$88.34_{\pm0.09}$  \\%
 
 Baseline(REF) &\pmb{$99.35_{\pm0.02}$} &$95.45_{\pm0.05}$ &$88.99_{\pm0.18}$ &$60.79_{\pm0.17}$ &$15.14_{\pm0.11}$ &$1.39_{\pm0.06}$ &$0.14_{\pm0.01}$\\%
 
 Baseline(PGD+REF) &\pmb{$99.35_{\pm0.03}$} &$95.39_{\pm0.17}$ &$89.98_{\pm0.13}$ &$66.31_{\pm0.24}$ &$22.75_{\pm0.21}$ &$2.58_{\pm0.14}$ &$0.36_{\pm0.04}$ \\%
 
 VIB-R &$99.10_{\pm0.05}$ &$92.06_{\pm0.19}$ &$43.28_{\pm0.24}$ &$2.55_{\pm0.08}$ &$0.03_{\pm0.02}$ &$0.00_{\pm0.00}$ &$0.00_{\pm0.00}$ \\ %
 
 VIB-M &$99.17_{\pm0.03}$ &$93.39_{\pm0.08}$ &$61.25_{\pm0.19}$ &$8.81_{\pm0.11}$ &$0.03_{\pm0.01}$ &$0.00_{\pm0.00}$ &$0.00_{\pm0.00}$ \\%
 
VIB-R(REF) &$99.24_{\pm0.03}$ &$95.19_{\pm0.07}$ &$86.33_{\pm0.24}$ &$55.69_{\pm0.50}$ &$17.51_{\pm0.23}$ &$2.15_{\pm0.10}$ &$0.27_{\pm0.03}$ \\ %
 
VIB-M(REF) &$99.28_{\pm0.02}$ &$95.17_{\pm0.06}$ &$86.28_{\pm0.23}$ &$53.82_{\pm0.32}$ &$13.53_{\pm0.05}$ &$1.64_{\pm0.09}$ &$0.19_{\pm0.02}$ \\%
 
VIB-M(PGD) &$99.04_{\pm0.05}$ &$98.26_{\pm0.02}$ &$97.19_{\pm0.06}$ &$95.84_{\pm0.06}$ &$94.03_{\pm0.13}$ &$90.86_{\pm0.14}$ &$85.51_{\pm0.18}$ \\ %
 
VIB-M(REF+FGSM) &$99.22_{\pm0.06}$ &$97.41_{\pm0.12}$ &$96.79_{\pm0.07}$ &$95.99_{\pm0.10}$ &$95.03_{\pm0.09}$ &$92.99_{\pm0.27}$ &$86.95_{\pm0.16}$ \\%
  
VIB-M(REF+PGD) &$99.13_{\pm0.04}$ &$98.24_{\pm0.06}$ &\pmb{$97.70_{\pm0.08}$} &\pmb{$97.36_{\pm0.08}$} &\pmb{$96.87_{\pm0.09}$} &\pmb{$96.01_{\pm0.06}$} &\pmb{$93.64_{\pm0.30}$} \\
 
 \hline
\end{tabular}
\end{table*}




\begin{table*}[ht!]
\small
\caption{Models' classification test accuracy results ($\%$) with the FGSM attack accuracy on the CIFAR-10 dataset.}
\label{Tab:fgsm_cifar10}
\centering
\begin{tabular}{ |l|c|c|c|c|c|c|c|} \hline
 Method & Clean & $\epsilon=0.01$ & $\epsilon=0.02$ & $\epsilon=0.03$ & $\epsilon=0.04$ & $\epsilon=0.05$ & $\epsilon=0.06$ \\ \hline
 
 Baseline &$92.88_{\pm0.13}$ &$86.36_{\pm0.14}$ &$83.67_{\pm0.18}$ &$81.68_{\pm0.23}$ &$79.99_{\pm0.18}$ &$78.52_{\pm0.14}$ &$76.88_{\pm0.41}$ \\%
 
 Baseline(KD) &\pmb{$93.41_{\pm0.06}$} &$84.34_{\pm0.24}$ &$80.42_{\pm0.20}$ &$77.58_{\pm0.29}$ &$75.44_{\pm0.30}$ &$73.56_{\pm0.19}$ &$71.74_{\pm0.64}$ \\%
 
 Baseline(PGD) &$90.24_{\pm0.19}$ &$84.71_{\pm0.22}$ &$79.36_{\pm0.31}$ &$75.49_{\pm0.21}$ &$72.72_{\pm0.11}$ &$70.48_{\pm0.17}$ &$68.99_{\pm0.14}$ \\%
 
 Baseline(REF) &$90.51_{\pm0.22}$ &$84.71_{\pm0.34}$ &$81.81_{\pm0.21}$ &$79.52_{\pm0.17}$ &$77.31_{\pm0.26}$ &$75.15_{\pm0.29}$ &$73.63_{\pm0.22}$\\  %
 
 Baseline(PGD+REF) &$87.11_{\pm0.08}$ &$83.05_{\pm0.21}$ &$79.29_{\pm0.04}$ &$76.40_{\pm0.17}$ &$73.91_{\pm0.22}$ &$71.77_{\pm0.14}$ &$69.76_{\pm0.22}$ \\%
 
 VIB-R &$90.57_{\pm0.07}$ &$84.66_{\pm0.32}$ &$81.24_{\pm0.14}$ &$78.66_{\pm0.27}$ &$76.47_{\pm0.19}$ &$74.84_{\pm0.25}$ &$73.03_{\pm0.17}$ \\ %
 
 VIB-M &$91.47_{\pm0.14}$ &$84.79_{\pm0.24}$ &$81.94_{\pm0.15}$ &$79.98_{\pm0.22}$ &$78.16_{\pm0.13}$ &$76.52_{\pm0.11}$ &$75.26_{\pm0.16}$ \\%
 
 VIB-R(REF) &$90.98_{\pm0.07}$ &$86.82_{\pm0.16}$ &$84.80_{\pm0.23}$ &$83.09_{\pm0.19}$ &$81.62_{\pm0.20}$ &$80.37_{\pm0.20}$ &$79.44_{\pm0.21}$ \\ %
 
 VIB-M(REF) &$91.49_{\pm0.05}$ &\pmb{$87.42_{\pm0.17}$} &\pmb{$85.32_{\pm0.13}$} &\pmb{$84.01_{\pm0.32}$} &\pmb{$82.42_{\pm0.26}$} &\pmb{$80.97_{\pm0.24}$} &\pmb{$80.05_{\pm0.42}$} \\ %
 
 VIB-M(PGD) &$74.55_{\pm0.03}$ &$69.43_{\pm0.26}$ &$63.74_{\pm0.26}$ &$58.88_{\pm0.28}$ &$53.87_{\pm0.31}$ &$49.67_{\pm0.28}$ &$46.10_{\pm0.30}$ \\ %
  
 VIB-M(REF+FGSM) &$87.29_{\pm0.09}$ &$83.62_{\pm0.19}$ &$81.56_{\pm0.18}$ &$80.11_{\pm0.09}$ &$78.61_{\pm0.16}$ &$77.48_{\pm0.18}$ &$76.17_{\pm0.26}$ \\%

 VIB-M(REF+PGD) &$89.57_{\pm0.16}$ &$86.24_{\pm0.13}$ &$83.69_{\pm0.18}$ &$82.06_{\pm0.17}$ &$80.23_{\pm0.36}$ &$78.78_{\pm0.30}$ &$77.52_{\pm0.18}$ \\%
 
\hline
\end{tabular}
\end{table*}


\begin{table*}[ht!]
\small
\caption{Models' classification test accuracy results ($\%$) with the PGD attack accuracy on the CIFAR-10 dataset.}
\label{Tab:pgd_cifar10}
\centering
\begin{tabular}{ |l|c|c|c|c|c|c|c|} \hline
 Method & Clean & $\epsilon=0.01$ & $\epsilon=0.02$ & $\epsilon=0.03$ & $\epsilon=0.04$ & $\epsilon=0.05$ & $\epsilon=0.06$ \\ \hline
 
 Baseline &$92.88_{\pm0.13}$ &$77.28_{\pm0.18}$ &$63.96_{\pm0.26}$ &$46.42_{\pm0.28}$ &$27.18_{\pm0.25}$ &$15.49_{\pm0.38}$ &$8.92_{\pm0.16}$ \\%
 
 Baseline(KD) &\pmb{$93.41_{\pm0.06}$} &$76.23_{\pm0.13}$ &$51.26_{\pm0.09}$ &$32.97_{\pm0.21}$ &$24.64_{\pm0.19}$ &$20.23_{\pm0.23}$ &$17.15_{\pm0.33}$ \\ %
 
 Baseline(PGD) &$90.24_{\pm0.19}$ &$84.00_{\pm0.11}$ &$76.02_{\pm0.23}$ &$67.08_{\pm0.15}$ &$58.13_{\pm0.22}$ &$50.37_{\pm0.23}$ &$44.79_{\pm0.20}$ \\%
 
 Baseline(REF) &$90.51_{\pm0.22}$ &$71.88_{\pm0.32}$ &$57.66_{\pm0.40}$ &$41.69_{\pm0.26}$ &$28.29_{\pm0.17}$ &$19.80_{\pm0.22}$&$14.60_{\pm0.19}$\\%
 
 Baseline(PGD+REF) &$87.11_{\pm0.08}$ &$82.56_{\pm0.11}$ &$77.21_{\pm0.16}$ &$71.65_{\pm0.21}$ &$65.17_{\pm0.36}$ &$58.65_{\pm0.24}$ &$52.40_{\pm0.21}$ \\%
 
 VIB-R &$90.57_{\pm0.07}$ &$79.06_{\pm0.34}$ &$63.62_{\pm0.25}$ &$48.99_{\pm0.33}$ &$36.59_{\pm0.13}$ &$26.93_{\pm0.21}$ &$19.86_{\pm0.15}$ \\ %
 
 VIB-M &$91.47_{\pm0.14}$ &$79.61_{\pm0.15}$ &$68.18_{\pm0.29}$ &$62.47_{\pm0.41}$ &$57.98_{\pm0.10}$ &$52.23_{\pm0.16}$ &$46.08_{\pm0.69}$\\ %
 
 VIB-R(REF) &$90.98_{\pm0.07}$ &$79.87_{\pm0.06}$ &$69.35_{\pm0.30}$ &$62.58_{\pm0.23}$ &$56.57_{\pm0.15}$ &$51.90_{\pm0.44}$ &$47.70_{\pm0.18}$\\%
 
 VIB-M(REF) &$91.49_{\pm0.05}$ &$80.55_{\pm0.19}$ &$70.52_{\pm0.18}$ &$60.05_{\pm0.24}$ &$50.94_{\pm0.20}$ &$44.01_{\pm0.09}$ &$39.21_{\pm0.47}$ \\%
 
 VIB-M(PGD) &$74.55_{\pm0.03}$ &$68.74_{\pm0.12}$ &$62.92_{\pm0.27}$ &$56.32_{\pm0.37}$ &$50.29_{\pm0.19}$ &$44.46_{\pm0.32}$ &$38.94_{\pm0.18}$ \\%
  
 VIB-M(REF+FGSM) &$87.29_{\pm0.09}$ &$82.47_{\pm0.11}$ &$77.22_{\pm0.18}$ &$71.54_{\pm0.34}$ &$66.11_{\pm0.33}$ &$60.35_{\pm0.18}$ &$55.53_{\pm0.14}$ \\%
 
 VIB-M(REF+PGD) &$89.57_{\pm0.16}$ &\pmb{$85.65_{\pm0.08}$} &\pmb{$80.92_{\pm0.13}$} &\pmb{$75.78_{\pm0.20}$} &\pmb{$69.86_{\pm0.20}$} &\pmb{$63.89_{\pm0.20}$} &\pmb{$59.87_{\pm0.24}$} \\ %

 \hline
\end{tabular}
\end{table*}

\end{document}